\documentclass[lettersize,journal]{IEEEtran}
\usepackage{amsmath,amsfonts}
\usepackage{algorithmic}
\usepackage{algorithm}
\usepackage{array}
\usepackage[caption=false,font=normalsize,labelfont=sf,textfont=sf]{subfig}
\usepackage{textcomp}
\usepackage{stfloats}
\usepackage{booktabs}
\usepackage{url}
\usepackage{verbatim}
\usepackage{amsmath}
\usepackage{xcolor}
\usepackage{mathrsfs}
\usepackage{multirow}
\usepackage{graphicx}
\usepackage{cite}

\usepackage{makecell}

\usepackage{hyperref}  
\hypersetup{colorlinks=true, citecolor=blue}

\begin{document}

\title{Frequency Domain Unlocks New Perspectives for Abdominal Medical Image Segmentation}

% \author{IEEE Publication Technology,~\IEEEmembership{Staff,~IEEE,}}
%%        % <-this % stops a space

\author{Kai Han, Siqi Ma, Chengxuan Qian, Jun Chen, Chongwen Lyu, Yuqing Song, Zhe Liu
\thanks{This work was supported in part by the National Natural Science Foundation of China under Grant (62276116), Jiangsu Six Talent Peak Program under Grant (DZXX-122), Jiangsu Graduate Research Innovation Program under Grant (KY-CX23$\_$3677) and National Undergraduate Training Program on Innovation and Entrepreneurship (Grant No. 202410299049Z). (Corresponding authors: Zhe Liu.)}
\thanks{Kai Han, Siqi Ma, Jun Chen, Chongwen Lyu, Yuqing Song and Zhe Liu are with the School of Computer Science and Communication Engineering, Jiangsu University, Zhenjiang 212013, China (e-mail: 2112108003@stmail.ujs.edu.cn; 2212208031@stmail.ujs.edu.cn; chenjun@ujs.edu.cn; 2212308023@stmail.ujs.edu.cn; yqsong@ujs.edu.cn; 1000004088@ujs.edu.cn).}
\thanks{Chengxuan Qian is with the School of Mathematical Sciences, Jiangsu University, Zhenjiang 212013, China (e-mail: chengxuan.qian@stmail.ujs.edu.cn).}}

% \thanks{Manuscript received XX XX, Year; revised XX XX, Year.}

% The paper headers
\markboth{}%
{Shell \MakeLowercase{\textit{et al.}}: A Sample Article Using IEEEtran.cls for IEEE Journals}

%\IEEEpubid{0000--0000/00\$00.00~\copyright~2021 IEEE}
% Remember, if you use this you must call \IEEEpubidadjcol in the second
% column for its text to clear the IEEEpubid mark.

\maketitle

\begin{abstract}
Accurate segmentation of tumors and adjacent normal tissues in medical images is essential for surgical planning and tumor staging. Although foundation models generally perform well in segmentation tasks, they often struggle to focus on foreground areas in complex, low-contrast backgrounds, where some malignant tumors closely resemble normal organs, complicating contextual differentiation. To address these challenges, we propose the Foreground-Aware Spectrum Segmentation (FASS) framework. First, we introduce a foreground-aware module to amplify the distinction between background and the entire volume space, allowing the model to concentrate more effectively on target areas. Next, a feature-level frequency enhancement module, based on wavelet transform, extracts discriminative high-frequency features to enhance boundary recognition and detail perception. Eventually, we introduce an edge constraint module to preserve geometric continuity in segmentation boundaries. Extensive experiments on multiple medical datasets demonstrate superior performance across all metrics, validating the effectiveness of our framework, particularly in robustness under complex conditions and fine structure recognition. Our framework significantly enhances segmentation of low-contrast images, paving the way for applications in more diverse and complex medical imaging scenarios.
\end{abstract}


\section{Introduction}
\begin{figure}[!t]
	\centering
	\includegraphics[width=0.95\columnwidth]{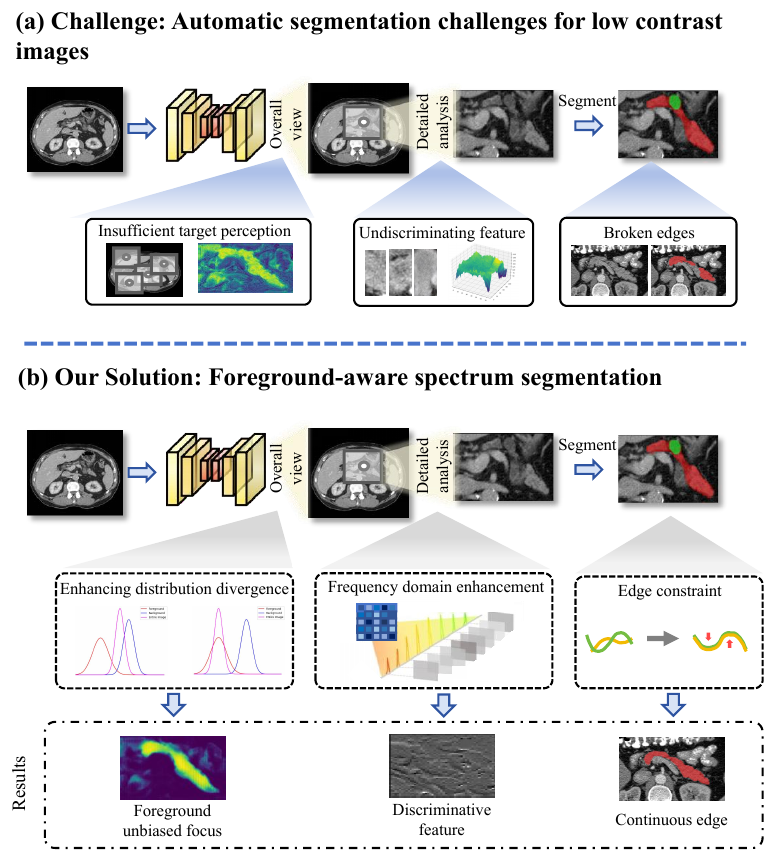}	
     \caption{Comparison between our FASS framework and previous automatic segmentation methods. (a) Segmentation process of low-contrast images by previous methods, which face challenges such as insufficient target perception, undiscriminating features, and broken edges. (b) Segmentation process of the FASS method. Our FASS framework employs adversarial training between the full image and background feature distribution to achieve focused attention on the foreground. Discriminative features are then enhanced in the frequency domain, and boundary integrity and continuity are strengthened through the edge constraint module.}	
	\label{fig1}
\end{figure}

% \IEEEPARstart{T}{he} identification of abdominal tumors is critical for early cancer detection, effective treatment planning, and improved patient survival rates \cite{siegel2024cancer,han2025climd,han2025region,qian2025adaptive}. By precisely segmenting organs and tumors in CT images, doctors can evaluate the condition of abdominal organs and the size, location, and shape of tumors, leading to a more accurate assessment of disease severity and the formulation of suitable treatment options \cite{liu20163d,dong2020weakly,shi2022polyp}. However, manual annotation is time-consuming, labor-intensive, and relies heavily on the expertise and experience of clinicians, especially for complex structures or unclear boundaries \cite{han2024deep,wang2023sac}. Consequently, the development of efficient and accurate automatic segmentation algorithms is of great importance.

\IEEEPARstart{T}{he} identification of abdominal tumors plays a pivotal role in early cancer detection, treatment planning, and improving patient survival \cite{siegel2024cancer,han2025climd,han2025region,qian2025adaptive}. Accurate segmentation of organs and tumors in CT images enables clinicians to assess organ conditions and precisely determine the size, location, and morphology of tumors, thereby facilitating more reliable disease evaluation and optimal therapeutic decision-making \cite{liu20163d,dong2020weakly,shi2022polyp}. However, manual annotation is extremely time-consuming and labor-intensive, and it demands substantial clinical expertise, particularly when dealing with complex anatomical structures or ambiguous boundaries \cite{han2024deep,wang2023sac}. Therefore, the development of efficient and accurate automated segmentation algorithms has become both essential and urgent.

These challenges are exacerbated in the segmentation of low-contrast abdominal images due to the intricate anatomical structures and limited contrast. In the abdominal region, multiple organs overlap, and tumors frequently adhere to or embed within organ surfaces, complicating foreground-background separation and leading to potential tissue misclassification. Additionally, the low contrast in these images results in blurred boundaries between tumors and surrounding tissues with similar grayscale values, making lesion contours challenging to discern. This difficulty highlights the need for advanced segmentation methods tailored to low-contrast environments.

Deep learning has shown promising potential in medical image segmentation, offering new approaches to address these complexities. Prior studies have employed two-stage strategies, progressing from manual region clipping \cite{zhou2019hyper} to model-driven autonomous learning \cite{chen2021model,ding2021tostagan,yu2020c2fnas,yuan2024sd,yuan2024tsar,yuan2025video,yuan2025msp,yuan2025sed,yuan2025msp,liu2024pancreas} for initial target localization. More recently, single-stage methods have emerged, enabling end-to-end segmentation of the target region \cite{jiang2022apaunet,li2023temperature}. In response to the challenges of low-contrast segmentation, approaches focused on boundary enhancement and multi-scale information fusion have been developed \cite{zhou2019high,zhang2024low}, aiming to improve segmentation by extracting high-resolution features and refining boundary details in blurred areas. Despite these advancements, current models continue to face limitations in low-contrast and complex abdominal environments. When target regions exhibit complex internal topologies and similar boundary pixel intensities, models often struggle to capture subtle discriminative features, resulting in incomplete foreground segmentation and broken boundaries. These challenges increase the difficulty of achieving precise segmentation, as illustrated in Fig.~\ref{fig1} (a).

To this end, we propose the Foreground-Aware Spectrum Segmentation (FASS) framework to improve target localization and detail capture, as illustrated in Fig.~\ref{fig1} (b). Specifically, the Foreground-Aware (FA) module enhances foreground feature extraction and interpretation by employing an adversarial training strategy to maximize the distributional contrast between background and input image features. After identifying the target region, we designed a Feature-Level Frequency Enhancement (FLFE) module to extract discriminative features. This module performs spectral decomposition of the encoded output using wavelet transform and enhances the complementarity of high-frequency details through a cross-attention mechanism. Subsequently, it selects high-discriminative high-frequency features using channel and spatial attention mechanisms, improving the perception of the target boundaries and internal structures. Additionally, we introduce an Edge Constraint (EC) module to ensure edge integrity and geometric continuity in the segmentation results. To sum up, our main contributions are as follows:
\begin{itemize}
	\item{We propose an end-to-end Foreground-Aware Spectrum Segmentation (FASS) framework tailored for low-contrast medical image segmentation tasks.} 
	\item{We design a Foreground-Aware (FA) module to deepen the model's understanding of foreground features by learning the heterogeneity between background and complete features, enabling focused attention on foreground regions.}
	\item{We introduce a Feature-Level Frequency Enhancement (FLFE) module based on wavelet transform, which selects discriminative high-frequency features to enhance detail capture.}
	\item{The Edge Constraint (EC) strategy is introduced to ensure boundary integrity and continuity, effectively preventing segmentation breaks in low-contrast settings.}
	\item{Extensive experiments demonstrate the independent performance benefits of each module within the FASS framework across multiple medical datasets, with overall performance significantly surpassing current state-of-the-art methods.}
\end{itemize}

\section{Related works}
\subsection{Low-Contrast Medical Image Segmentation}
Low-contrast medical image segmentation is a challenging task, especially when dealing with abdominal images from modalities like CT, MRI, or ultrasound. Automatic segmentation of such images is very difficult due to the small grayscale difference between the target and the background tissue. Traditional threshold segmentation methods or edge detection algorithms do not perform well in this context because they are highly dependent on sharp contrast differences. In recent years, deep learning has become a powerful tool in the field of medical image segmentation by learning complex features from images \cite{cciccek20163d,milletari2016v,chen20233d,isensee2021nnu}. However, due to the subtle differences between the foreground and background, deep learning models can still encounter challenges in accurately distinguishing the boundaries of target regions in certain cases. To this end, researchers utilize generative adversarial networks (GANs) or enhancement techniques to generate clearer low-contrast images to assist in segmentation tasks \cite{subramani2023optimal,xu2024deep}. 
However, such methods often rely on the diversity of the original data, which may introduce artifacts or unrealistic features, affecting the reliability of the segmentation results. Besides, integrating information from other modalities/centers can compensate for the limitations of a single modality/center in low-contrast scenarios \cite{xiang2024unpaired,fu2024generalizable}, but acquiring such data is costly and requires substantial computational resources. There are also methods that attempt to enhance features or information within the network to improve the segmentation performance of low-contrast images \cite{zhou2019high,zhang2024low}. Despite some progress, low-contrast images typically exhibit minimal differences between the foreground and background, making it difficult for spatial domain enhancement to significantly improve these subtle distinctions. Our method utilizes frequency domain enhancement to amplify high-frequency components, enabling the model to better capture fine structures and accurately segment subtle features in complex scenes.

\subsection{Region of Interest Location}
The inherent complexity of medical images, particularly the similarity in texture, brightness, and morphology between background and foreground, poses a significant challenge to the localization of the region of interest. Traditional methods rely on manually segmenting the foreground region, which is effective but limited by the dependence on expert knowledge and the lack of automation \cite{zhou2019hyper,han2024limt,yuan2025dvp,yuan2025dvplus,chen2025haif,zhu2025contrastive,li2025rusc}, making them unsuitable for large-scale image analysis. In recent years, two-stage segmentation methods have somewhat alleviated the problem of background interference, but they have introduced increased algorithmic complexity and the risk of error accumulation \cite{chen2021model,ding2021tostagan,yu2020c2fnas,liu2024pancreas}. Against this backdrop, single-stage region-of-interest segmentation models have emerged. For example, Jiang et al. \cite{jiang2022apaunet} proposed the axial projection attention unit, which effectively filters out redundant feature information. Li et al. \cite{li2023temperature} introduced a balanced temperature loss function, significantly enhancing the model's focus on target regions. Besides, multi-task learning frameworks \cite{chen2019multi,qian2025decalign,qian2025dyncim,hao2024ssdc,yuan2025autodrive,yuan2025video,xu2023pac} have been used to simultaneously predict multiple related outputs, enhancing the understanding of foreground details. Attention mechanisms \cite{wang2021self,valanarasu2021medical,yu2023adaptive} have also been introduced, allowing the network to adjust its focus on different regions of the image, thus concentrating on key foreground structures. Despite these advances, these methods learn an unbiased mixture of foreground and background features during training, limiting the model's ability to deeply explore their differences. By comparison, our method ensures that the model can pay biased attention to the region of interest during the inference stage, demonstrating excellent adaptability even in low-contrast environments by effectively resisting complex background interference.

\subsection{Frequency-Based Image Analysis Techniques}
Despite significant advances in image segmentation achieved through deep learning techniques, existing methods often rely on simulating human visual perception processes. These methods tend to integrate and process high-frequency (e.g., edges and textures) and low-frequency (e.g., shapes) information at the visual level. However, this practice may face limitations in low-contrast conditions due to difficulties in precisely distinguishing subtle discriminative features. Wavelet transform, as a tool with excellent spatial representation capabilities and directional sensitivity, can decompose image features into different frequency components, offering a method to fully utilize frequency information \cite{tian2023multi,yin2021method}. 

In the field of medical image segmentation, high-frequency features extracted using wavelet transform have been proven to significantly enhance neural networks' ability to learn high-frequency details \cite{zhao2023wranet,gao2022medical,showrav2024hi}. By capturing details that are easily overlooked by human vision, networks can effectively address the challenges of segmentation under low-contrast conditions. Based on this, Jin et al. \cite{jin2023novel} explored frequency feature fusion techniques to improve models' grasp of detailed textures and overall structures. Azad et al. \cite{azad2021deep}, on the other hand, advocated for moderately suppressing high-frequency information to reduce excessive reliance on texture details, though this strategy may fall short in a low-contrast environment. Although the above strategies have achieved important breakthroughs in utilizing frequency information, indiscriminate use of high-frequency features may introduce noise, potentially threatening segmentation accuracy. In light of this, this paper proposes an effective strategy to enhance and selectively utilize discriminative high-frequency information to improve model performance under low-contrast conditions.

\section{Method}

\begin{figure*}[!t]
\centering
\includegraphics[width=0.95\linewidth]{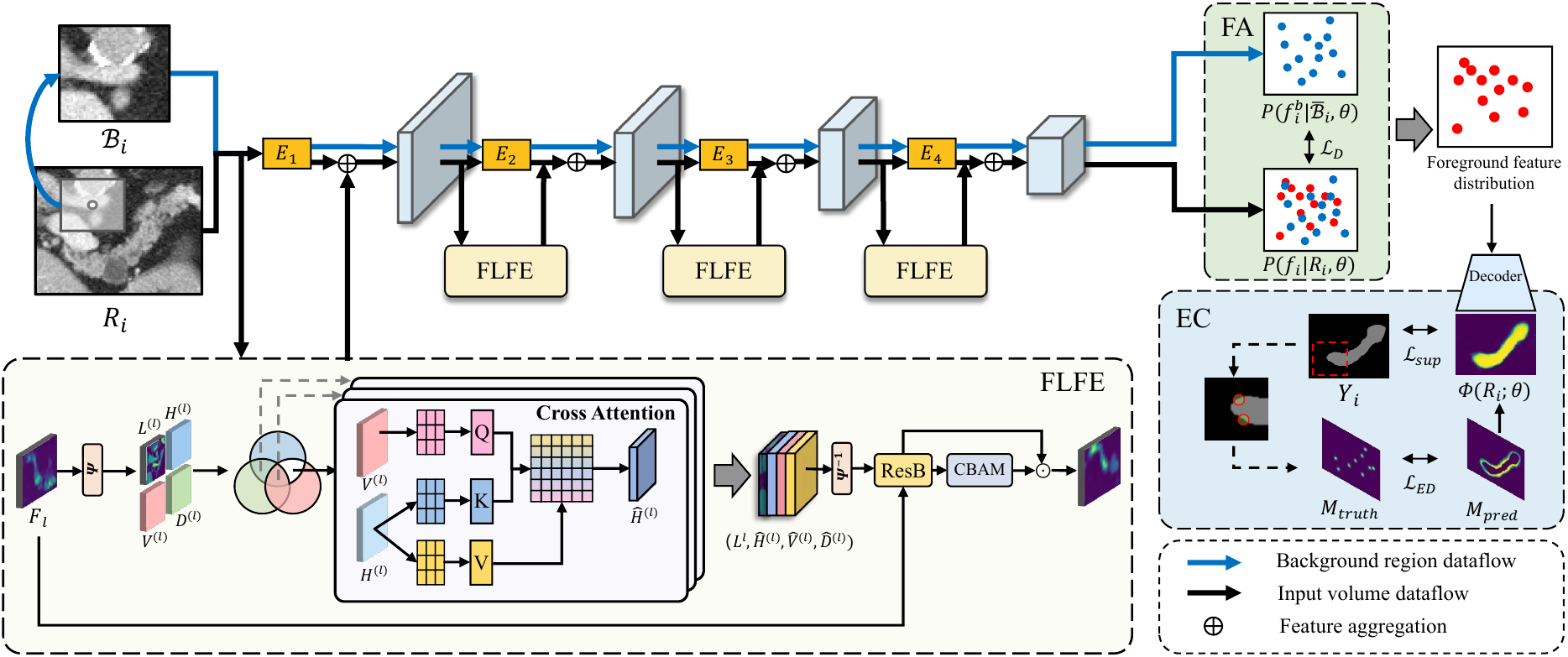}
  \caption{Overview of the proposed FASS framework. The framework consists of a foreground-aware (FA) module (Sec. \ref{sec3.1}), a feature-level frequency enhancement (FLFE) module (Sec. \ref{sec3.2}), and an edge constraint (EC) module (Sec. \ref{sec3.3}). Initially, sampled patches and background patches are fed into the encoder for feature extraction. The feature differences between encoder outputs are computed and maximized, with the FLFE module enhancing features during the encoding phase. Finally, the EC module refines the edges of the decoder output for optimized segmentation results.}
	\label{fig2}
\end{figure*}

In this section, we present our Foreground-Aware Spectrum Segmentation (FASS) framework, which consists of three key modules: the Foreground-Aware (FA) module, the Feature-Level Frequency Enhancement (FLFE) module, and the Edge Constraint (EC) module. The FA module employs adversarial training to maximize the feature distribution differences between the background and the input volume, guiding the model to focus more effectively on the foreground. The FLFE module selects more discriminative high-frequency features to enhance the ability to capture details. Furthermore, the EC module refines the morphological contours of edge predictions to further enhance segmentation accuracy. The overall framework is illustrated in Fig.~\ref{fig2}.

\subsection{Foreground-Aware Module}\label{sec3.1}
To address the issue of the minimal difference between foreground and background in low-contrast images, which makes it difficult for the model to accurately identify the foreground, we introduce a Foreground-Aware (FA) module. Suppose an input image patch space $\{R_i\in I_i,i\in N\}$ with labels $\{Y_{R_i},i\in N\}$ random cropped from corresponding image volume $I_i\in{\mathbb{R}^{w'\times h'\times d'}}$, where ${w'}\times{h'}\times{d'}$ represent the dimension of the volume, $N$ is the sample number, and $Y_i$ is the corresponding label. In this section, our goal is to randomly sample a background region $\mathcal{B}_i$ of size ${w}\times{h}\times{d}$ from $R_i$, as described in Eq.~\ref{eq1}:
\begin{equation}\label{eq1}
\begin{aligned}
		 \mathcal{B}_i &= R_i[x:x+w,y:y+h,z:z+d], \\&x\sim{Uniform}(0,w'-w),\\
		 &y\sim{Uniform}(0,h'-h),\\
		 &z\sim{Uniform}(0,d'-d).
\end{aligned}
\end{equation}
where $x$, $y$ and $z$ represent the random coordinate position of $R_i$. $Uniform(\cdot,\cdot)$ represents the randomness of the position point value.

In order to ensure that the sampled background region contains discrimination features from the foreground region, a parameter $\alpha$ is introduced as a control parameter for background region sampling to quantify the overlap degree between the background and the foreground region. Only if the overlap volume between is less than $\alpha$, the background region $\bar{\mathcal{B}_i}$ will be selected for training. The above process can be expressed as:
\begin{equation}\label{eq2}
\bar{\mathcal{B}_i}={\mathcal{B}_{i}\in{R_i}:Inter(\mathcal{B}_{i},\mathcal{F}_i)<\alpha}
\end{equation}
where $\mathcal{F}_i\in{R_i}$ is the foreground region and ${Inter}(\cdot,\cdot)$ represents the intersection calculation of two sets and is defined as follow:
\begin{equation}\label{eq3}
	Inter(\mathcal{B}_{i},\mathcal{F}_i)=|\mathcal{B}_{i}\bigcap{\mathcal{F}_i}|
\end{equation}
The selection of background region sampling size and hyperparameter $\alpha$ will be discussed in detail in Sec. \ref{sec4.5}.

Then, the dual-path encoder architecture is adopted to capture the background region feature ${f^b_i}$ and the global feature ${f_i}$, respectively. The encoder architecture consists of four layers of multi-scale convolutions. In order to further optimize the feature representation, a distribution divergence loss $\ell_{KL}$ based on KL divergence is introduced to minimize the distribution divergence $(min|P(f_i|R_i,\theta)-P(f^b_i|{\bar{\mathcal{B}_i}},\theta)|)$. The distribution difference loss $\ell_{KL}$ can be expressed as:
\begin{equation}\label{eq4}
    \begin{aligned}
    \textbf{D}(P(f_i|R_i,\theta)&\Vert{P(f^b_i|\bar{\mathcal{B}_i},\theta)}) 
     \\
    &=\quad P(f_i|R_i,\theta)\log\frac{P(f_i|R_i,\theta)}{P(f^b_i|\bar{\mathcal{B}_i},\theta)}
    \end{aligned}
\end{equation}

\begin{equation}\label{eq5}
	\ell_{KL} = e^{-\textbf{D}(P(f_i|R_i,\theta)\Vert{P(f^b_i|{\bar{\mathcal{B}_i}},\theta)})}
\end{equation}
where $\theta$ represents the parameters of the encoder.

In addition, the random cropping strategy of the input image $I_i$ may lead to variable proportion of the foreground region $\mathcal{F}_i$ in the image patch ${R_i}$, which has a significant impact on the learning effect of this module. A higher proportion of foreground region is conducive to the model learning how to accurately focus on the foreground region, also meaning that the patch ${R_i}$ is more valuable for training. In view of this, we design an adaptive weight factor ${\omega}$ to ensure that the model can learn from images with different foreground proportions in a balanced way by adjusting the loss contribution of each sampled patch ${R_i}$ for better understanding the essential differences between foreground and background features, as shown in Eq.~\ref{eq6}:
\begin{equation}\label{eq6}
\omega = Inter(I_i^{fore},\mathcal{F}_i)
\end{equation}
where $I_i^{fore}$ is the volume of the foreground region in global image volume $I_i$. The total loss function of this module can be expressed as follows:
\begin{equation}\label{eq7}
	\mathop{\min}_{\theta} \mathcal{L}_D = \omega\cdot\mathop{\min\ell_{KL}}_{\theta}
\end{equation}

\subsection{Feature-Level Frequency Enhancement Module} \label{sec3.2}
Suppose ${F_l}$ denotes the feature map of encoder layer ${l}$. To enhance the ability to capture details using frequency information, the wavelet transform $\Psi(\cdot ,\varepsilon_w)$ is used to map the spatial domain $Y_i$ to the frequency domain $Y^{(i)}$ $(e.g., F_l\rightarrow F^{(l)})$, where $\varepsilon_w$ represents different wavelet bases. Specifically, it applies the low-pass filter ${f_g}$ and the high-pass filter ${f_h}$, and the subsequent down-sampling operation $\downarrow_2$ on ${F^{l}}$ to obtain a set of approximation coefficients ${(L^{(l)})}$ and detail coefficients ${(H^{(l)},V^{(l)},D^{(l)})}$, which correspond to the low frequency (overall structure) and high frequency (texture and edge) information, respectively. The above process can be expressed as follows:
\begin{equation}\label{eq8}
	\begin{aligned}
			L^{(l)},H^{(l)},V^{(l)},D^{(l)} = & \downarrow_2(f_g\ast{F_l}),\downarrow_2(f_h\ast{F_{l}}),\\
			&\downarrow_2(f_h\ast{F_{l}^{T}}),\downarrow_2(f_h\ast{f_h^T}\ast{F_l})
	\end{aligned}
\end{equation}
where $\ast$, ${\downarrow_2}$, $T$ denote convolution operation, down-sampling operation, and transpose operation, respectively.

To further enhance the detail richness in the high-frequency components, a cross-attention mechanism is introduced. This mechanism promotes the model to learn from each other and construct a more comprehensive feature representation. Here, taking the high-frequency components ${H^{(l)}}$ and ${V^{(l)}}$ as an example, the process can be expressed as:
\begin{equation}\label{eq9}
	\footnotesize
\hat{H}^{(l)} = \sigma(W_H(H^{(l)}\cdot{A_V V^{(l)}})) 
\end{equation}
\begin{equation}\label{eq10}
	\footnotesize
 \hat{V}^{(l)} = \sigma(W_V(V^{(l)}\cdot{A_H H^{(l)}}))
\end{equation}
where ${\hat{H}}$ and ${\hat{V}}$ represent the horizontal and vertical high-frequency components after cross-attention fusion, respectively. ${\sigma}$ is the activation function, and ${W_H}$ and ${W_V}$ are the weight matrices used to adjust the feature fusion.

The attention weight matrix $A_V$ and $A_H$ are calculated based on the correlation between high-frequency components, which can be expressed as:
\begin{equation}\label{eq11}
	\footnotesize
A_H = softmax(\frac{Q_{\hat{H}^{(l)}}K_{\hat{H}^{(l)}}^T}{\sqrt{d}}),A_V = softmax(\frac{Q_{\hat{V}^{(l)}}K_{\hat{V}^{(l)}}^T}{\sqrt{d}})
\end{equation}
where ${Q_{H^{(l)}}}$ and ${K_{H^{(l)}}}$ denote the query and key matrices of ${H^{(l)}}$, respectively. The ${softmax}$ function is used to normalize the attention weights; ${d}$ is the dimension of the key vector used to scale the inner product.

After the above steps, the enhanced high-frequency components ${(\bar{H}^{(l)},\bar{V}^{(l)},\bar{D}^{(l)})}$ are obtained. In order to reconstruct back to the spatial domain, the inverse wavelet transform is applied to obtain the enhanced feature map. The process is shown in Eq.~\ref{eq12}:
\begin{equation}\label{eq12}
F_{l}^{'} = \Psi^{-1}(L^l,\bar{H}^{(l)},\bar{V}^{(l)},\bar{D}^{(l)})
\end{equation}
where $\Psi^{-1}{(\cdot,\varepsilon_w)}$ denotes the inverse wavelet transform operation. Subsequently, in order to preserve long-range dependencies, residual block $Res(\cdot)$ with batch normalization is used. We further introduce a $CBAM(\cdot)$ module \cite{woo2018cbam} to ensure the model focuses on high-frequency information that is critical to the task. The generated attention map $P_{l}$ can be expressed as:
\begin{equation}\label{eq13}
	P_{l}=CBAM(Res(F^{'}_l))
\end{equation}

In view of the structural characteristics of the U-shaped network, its shallow layers tend to capture high-frequency detailed information, while the deep layers focus more on low-frequency global semantic features. Therefore, we gradually aggregate $F_l^{'}$ into deeper network layers in the encoders so as to ensure feature representation retains detail richness and semantic understanding ability. The aggregated feature map $F_{l+1}^{agg}$ can be expressed as:
\begin{equation}\label{eq14}
	F_{l+1}^{agg}=FA_{E}(F_{l+1},(F_l^{'}\odot{P_{l}}))
\end{equation}
where $FA_{E}(\cdot)$ represents the feature aggregation of the encoder layers, and $\odot$ represents Hadamard product.

\subsection{Edge Constraint Module}\label{sec3.3}
In order to further improve the integrity and the continuity of the geometric shape under the condition of low contrast, the Edge Constraint (EC) module is introduced. The module integrates the prior knowledge of the physical model into the deep learning framework to generate segmentation results that are more in line with the ground truth geometry.

Specifically, an initial set of boundary points ${B}$ is first extracted from the input image with the help of traditional edge detection algorithms. Then, define a circular window centered $O(r,b_i)$ at the boundary point $b_i\in B$ with a radius $r$ of (10 pixels by default), and calculate the proportion $p(b_i)$ of the foreground area within this window. As shown in Eq.~\ref{eq15}:
\begin{equation}\label{eq15}
p(b_i)=\frac{Inter(O(r,b_i),I^{fore})}{O(r,b_i)}
\end{equation}

When the ratio $p(b_i)$ is closer to 0 or 1, it indicates irregular boundaries within that window. To quantify this irregularity, a scoring function $s(b_i)$ is introduced as shown in Eq.~\ref{eq16}:
\begin{equation}\label{eq16}
	s(b_i)=|p(b_i)-0.5|
\end{equation}

The higher the score function $s(b_i)$, the greater the irregular near the boundary point $b_i$, and the more helpful it is for edge continuity learning. In order to highlight the key boundary features, the non-maximum suppression (NMS) technique is used to filter the local maximum of the scoring function $s(b_i)$. Specifically, for each boundary point $b_i$, we compare its scores with those of its $k$ nearest neighbors ($k$ is set to 10 by default) and keep only those boundary points whose scores are greater than nearest neighbors. Based on the filtered set of boundary points, the label of the size region around the retained boundary point $b_i$ is set to 1, while other points are set to 0. Thus, a ground truth map of the boundary key point set $M_{truth}$ is obtained. Different from $M_{truth}$, the predicted boundary key point set $M_{pred}$ is retained as the predicted probability value.

By imposing constraints on the key point set $M_{pred}$ predicted by the network and the ground truth key point set $M_{truth}$, the network is guided to generate a more complete and continuous boundary representation. The loss $\mathcal{L}_{match}$ between $M_{pred}$ and $M_{truth}$ is measured in the form of a cross-entropy loss function $\mathcal{L}_{CE}$, which is defined as:
\begin{equation}\label{eq17}
	\mathcal{L}_{match}=\mathcal{L}_{CE}(M_T,M_{pred})
\end{equation}

The boundary coherence loss $\mathcal{L}_{cont}$ focuses on the spatial relationship between key points in $M_{pred}$, ensuring that they form a continuous path on the image. This is achieved by calculating the predicted boundary key point difference between adjacent pixels, which is defined as follows:
\begin{equation}\label{eq18}
	\mathcal{L}_{cont} = \sum_{i-1}^{N-1}\rho_{i,i+1}\cdot|M_{pred}^i-M_{pred}^{i+1}|
\end{equation}
where $\rho_{i,i+1}$ is a weight calculated by the distance of $\lVert M_{pred}^i-M_{truth}^j\lVert_2-\lVert M_{pred}^{i+1}-M_{truth}^{j+1}\lVert_2$ and $\lVert M_{pred}^i-M_{pred}^{i+1}\lVert_2$, to reflect the coherence requirements between pixels, $j$ denotes the point nearst $i$. The total loss of the EC module is defined as:
\begin{equation}\label{eq21}
\mathcal{L}_{EC}=\frac{1}{2}\cdot(\mathcal{L}_{match}+\mathcal{L}_{cont})
\end{equation}

\subsection{Loss function}
Our FASS framework overall loss consists of three parts: the supervised loss $\mathcal{L}_{sup}$, the distribution difference loss $\mathcal{L}_D$, and the edge constraint loss $\mathcal{L}_{EC}$, as shown in Eq.~\ref{eq19}.
\begin{equation}\label{eq19}
	\mathcal{L}_{total} = \mathcal{L}_{sup}+ \lambda(t)(\mathcal{L}_D + \mathcal{L}_{EC})
\end{equation}
where $\lambda(t)$ is the time-varying Gaussian heating coefficient \cite{adiga2024anatomically} used to balance the above loss, which can be expressed as $\lambda(t)=0.1 * \exp{(-5(1-\frac{t}{t_{max}})^2)}$, $t$ and $t_{max}$ denote the current iteration number and the total iteration number, respectively. It is worth noting that we adopt hybrid loss as the supervision loss $L_{sup}$, which can be expressed as:
\begin{equation}\label{eq20}
	\small
	\mathcal{L}_{sup}=\frac{1}{2}\cdot(\mathcal{L}_{Dice}(R_i,Y_{R_i})+\mathcal{L}_{CE}(R_i,Y_{R_i}))
\end{equation}

\section{Experiment}
\subsection{Datasets}
To comprehensively evaluate our algorithm, we selected three representative low-contrast abdominal image datasets.

\noindent\textbf{MSD Pancreas Dataset} provided by the MICCAI 2018 Medical Segmentation Decathlon (MSD) challenge \cite{simpson2019large}. It comprises 281 CT scans along with their corresponding pancreas and tumor labels, with a median resolution of $0.8\times 0.8 \times 2.5$ mm$^{3}$. Out of these, 225 scans were used for training, while the rest were reserved for testing.

\noindent\textbf{NIH Dataset} consists of 82 abdominal CT scans annotated with pancreas labels \cite{roth2015deeporgan}. Volume sizes range from $512 \times 512 \times 181$ to $512 \times 512 \times 466$. Following \cite{liu2024pancreas}, we selected 62 samples for training and reserved 20 samples for testing.

\noindent\textbf{LiMT Dataset} were collected from the Affiliated Hospital of Jiangsu University. It covers four types of liver diseases: hepatocellular carcinoma (HCC), metastatic liver cancer, hemangioma, and liver cyst. The dataset includes 100 volumes of arterial phase CT scans, each annotated and verified by experienced clinical experts. For the experiments, 80 scans were used for training, while the remaining 20 were used for testing.

The MSD pancreas dataset is used to highlight the relationship between organs and small tumors; the NIH dataset evaluates segmentation performance for a single organ within complex backgrounds; and the LiMT dataset further assesses the model's ability to segment organs and distinguish various tumor subtypes.

\subsection{Evaluation Metrics}
To comprehensively evaluate the proposed segmentation method, we employ four metrics: Dice Similarity Coefficient (Dice), Jaccard Index (Jaccard), 95th Hausdorff Distance (95HD), and Average Surface Distance (ASD). The Dice and Jaccard metrics evaluate overlap with ground truth, where higher values indicate better overlap. 95HD measures boundary discrepancy, and ASD reflects average surface distance, with lower values indicating closer alignment. 

\subsection{Implementation Details}
All experiments were conducted on a computing platform with an NVIDIA RTX A6000 GPU and 48 GB of RAM. Given U-Net's \cite{cciccek20163d} strong performance and adaptability in medical image segmentation, we adopted it as our baseline network. Each experiment ran for 30,000 iterations to thoroughly optimize the model parameters. Model optimization was performed using Stochastic Gradient Descent (SGD) with an initial learning rate of 0.01, momentum of 0.9, and weight decay of 0.0001 to enhance generalization. Random rotation is introduced to do data augmentation. A five-fold cross-validation was employed for systematic evaluation. The selection of parameter $\alpha$ and its impact on model performance will be detailed discussed in the subsequent parameter sensitivity analysis section \ref{sec4.5}.

\subsection{Comparison with State-of-the-Art Methods}

To demonstrate the superiority of our framework, we conducted comprehensive evaluations across three medical datasets to systematically compare our approach against methods from three different research directions. These directions cover: i) \textit{Mainstream medical image segmentation methods:} 3D U-Net \cite{cciccek20163d}, V-Net \cite{milletari2016v}, Swin UNETR \cite{tang2022self}, RC-3DUNet \cite{liu2024pancreas}, nnU-Net (3D) \cite{isensee2021nnu}, and U-Mamba \cite{ma2024u}; ii) \textit{Frequency domain-enhanced segmentation methods:} WU-Net \cite{li2020wavelet}, XNet \cite{zhou2023xnet}, FET \cite{azad2023unlocking}, and SASAN \cite{huang2024sasan}; iii) \textit{Low-contrast medical image segmentation methods:} HMEDN \cite{zhou2019high} and TBNet \cite{zhang2024low}.

\begin{table*}
	\begin{center}
		\caption{Segmentation results on the MSD dataset compared with other state-of-the-art approaches.}
		\label{tab1}
		\resizebox{\linewidth}{!}{
			\begin{tabular}{rccccccccccc}
				\toprule
				\multirow{3}{*}{Methods}&\multicolumn{11}{c}{Metrics}\\
				\cline{2-12}
				&\multicolumn{2}{c}{Dice [\%]$\uparrow$}&&\multicolumn{2}{c}{Jaccard [\%]$\uparrow$}&&\multicolumn{2}{c}{95HD [mm]$\downarrow$}&&\multicolumn{2}{c}{ASD [mm]$\downarrow$}\\
				\cline{2-3}\cline{5-6}\cline{8-9}\cline{11-12}
				&Pancreas&Tumor&&Pancreas& Tumor &&Pancreas&Tumor &&Pancreas&Tumor\\
				\midrule
				3D U-Net \cite{cciccek20163d}       &79.52$\pm$7.54  &43.18$\pm$28.50  &&65.96$\pm$5.32  &31.53$\pm$20.51 &&7.69$\pm$0.83  &35.49$\pm$8.39 && 3.03$\pm$0.56 &12.20$\pm$2.39\\
				V-Net \cite{milletari2016v}         &78.19$\pm$8.45  &41.75$\pm$30.49  &&64.75$\pm$6.51  &29.38$\pm$24.55 &&8.00$\pm$1.23  &37.74$\pm$8.67 && 2.95$\pm$0.64 &13.30$\pm$3.64\\
                    Swin UNETR \cite{tang2022self}      &79.71$\pm$8.61  &40.83$\pm$26.78  &&66.28$\pm$7.87  &26.15$\pm$22.57 &&5.86$\pm$0.76  &38.70$\pm$9.73 && 2.71$\pm$0.27 &10.88$\pm$3.83\\
				RC-3DUNet \cite{liu2024pancreas}   &84.83$\pm$6.55  &48.36$\pm$26.50  &&73.25$\pm$6.94  &42.99$\pm$17.83 &&4.44$\pm$0.67  &26.98$\pm$5.14 && 1.70$\pm$0.38 & 5.45$\pm$2.29\\
                    nnU-Net(3D) \cite{isensee2021nnu}   &84.24$\pm$7.53  &49.83$\pm$25.14  &&72.78$\pm$5.17  &42.83$\pm$16.75 &&4.22$\pm$0.49  &27.07$\pm$4.67 && 1.94$\pm$0.37 & 8.47$\pm$1.30\\
				U-Mamba \cite{ma2024u}              &84.85$\pm$6.01  &58.55$\pm$27.28  &&73.69$\pm$6.54  &32.06$\pm$19.30 &&3.73$\pm$0.39  &26.74$\pm$5.28 && 1.35$\pm$0.42 & 5.83$\pm$1.20\\
    		\midrule
				WU-Net \cite{li2020wavelet}         &81.95$\pm$8.08  &51.73$\pm$27.43  &&70.07$\pm$6.28  &34.85$\pm$21.82 &&4.54$\pm$0.45  &26.51$\pm$6.25 && 1.85$\pm$0.26 & 5.34$\pm$1.42\\
				XNet \cite{zhou2023xnet}            &85.73$\pm$7.41  &52.15$\pm$25.17  &&75.40$\pm$6.75  &35.50$\pm$18.55 &&3.82$\pm$0.32  &27.44$\pm$7.31 && 1.46$\pm$0.55 & 4.90$\pm$1.82\\
				FET \cite{azad2023unlocking}        &84.98$\pm$5.71  &57.38$\pm$23.34  &&73.95$\pm$5.81  &42.73$\pm$16.38 &&4.38$\pm$0.51  &24.46$\pm$5.23 && 1.28$\pm$0.35 & 5.74$\pm$1.29\\ 
    			SASAN \cite{huang2024sasan}         &85.29$\pm$5.96  &54.55$\pm$22.12  &&74.71$\pm$6.93  &37.42$\pm$20.72 && 5.49$\pm$0.78  &25.47$\pm$4.63 && 1.99$\pm$0.48 & 5.58$\pm$1.55\\
    		\midrule    
				HMEDN \cite{zhou2019high}           &83.95$\pm$8.64  &50.37$\pm$24.88  &&72.59$\pm$6.76  &33.54$\pm$26.43 && 5.76$\pm$0.81  &26.09$\pm$5.52 && 1.53$\pm$0.27 & 6.32$\pm$1.47\\
				TBNet \cite{zhang2024low}           &84.73$\pm$9.92  &57.08$\pm$21.78  &&73.68$\pm$7.53  &39.84$\pm$19.30 && 5.22$\pm$0.62  &25.70$\pm$5.42 && 1.15$\pm$0.29 & 5.13$\pm$1.42\\
    		\midrule
				Ours                                &\textbf{87.85$\pm$5.30} &\textbf{60.49$\pm$18.26}  &&\textbf{78.02$\pm$4.98}  &\textbf{43.58$\pm$15.30} && \textbf{3.55$\pm$0.29}  &\textbf{23.99$\pm$4.02} && \textbf{1.06$\pm$0.21} & \textbf{4.55$\pm$0.87}\\
				\bottomrule 
		\end{tabular}}
	\end{center}
\end{table*}

\subsubsection{Quantitative Analysis}
We report the comparison of our FASS with state-of-the-art segmentation methods on the MSD pancreas dataset, as shown in Table~\ref{tab1}. From this table, our FASS achieves a Dice score of 87.85\% and a Jaccard index of 78.02\% in pancreas segmentation, marking an improvement of at least 2\% over other general and frequency-domain methods. In tumor segmentation, our model reaches a Dice score of 60.49\%, which is at least 3\% higher than the best frequency-domain methods (FET and WU-Net), and shows improved overlap in the Jaccard index. In addition, our model achieves low boundary error values, with ASD scores of 1.06 mm for pancreas and 4.55 mm for tumor segmentation, both notably lower than those of other methods.

\begin{table}
	\begin{center}
		\caption{Segmentation results on NIH dataset compared with other state-of-the-art methods.}
		\label{tab2}
		\resizebox{\linewidth}{!}{
			\begin{tabular}{rccccccc}
				\toprule
				\multirow{2}{*}{Methods}&\multicolumn{6}{c}{Metrics}\\
				\cline{2-8}
				&Dice [\%]$\uparrow$&&Jaccard [\%]$\uparrow$&&95HD [mm]$\downarrow$&&ASD [mm]$\downarrow$\\
				\midrule
				3D U-Net \cite{cciccek20163d}      &79.48$\pm$5.06 && 66.07$\pm$4.64 &&  5.53$\pm$1.40 && 2.57$\pm$1.14\\
				V-Net \cite{milletari2016v}        &78.53$\pm$7.81 && 64.89$\pm$5.32 &&  5.83$\pm$1.23 && 2.66$\pm$0.93 \\
    			Swin UNETR \cite{tang2022self}     &80.74$\pm$6.23 && 67.78$\pm$5.41 &&  4.66$\pm$1.56 && 1.74$\pm$1.02\\
    			RC-3DUNet \cite{liu2024pancreas}  &84.62$\pm$4.85 && 74.51$\pm$1.92 &&  3.23$\pm$0.89 && 1.13$\pm$0.51\\
       		  nnU-Net(3D) \cite{isensee2021nnu}  &85.67$\pm$4.25 && 75.02$\pm$3.34 &&  3.36$\pm$0.57 && 1.12$\pm$0.67\\
				U-Mamba \cite{ma2024u}             &86.08$\pm$3.95 && 75.57$\pm$2.14 &&  3.29$\pm$0.53 && 0.93$\pm$0.37\\
        		\midrule
				XU-Net \cite{li2020wavelet}        &82.45$\pm$7.56 && 70.36$\pm$5.73 &&  4.25$\pm$1.26 && 1.07$\pm$0.52\\    
				XNet \cite{zhou2023xnet}           &85.67$\pm$8.01 && 74.97$\pm$4.19 &&  5.24$\pm$0.82 && 1.53$\pm$0.91\\
				FET \cite{azad2023unlocking}       &83.29$\pm$5.15 && 71.49$\pm$4.51 &&  3.73$\pm$0.97 && 1.20$\pm$0.73\\
                    SASAN \cite{huang2024sasan}        &85.30$\pm$3.99 && 74.39$\pm$5.73 &&  3.17$\pm$1.05 && 1.16$\pm$0.67\\
        		\midrule
          	  HMEDN \cite{zhou2019high}          &84.89$\pm$6.17 && 73.68$\pm$4.61 &&  3.20$\pm$0.72 && 1.28$\pm$0.88\\
                    TBNet \cite{zhang2024low}          &85.03$\pm$6.83 && 73.84$\pm$5.08 &&  2.97$\pm$0.83 && 1.03$\pm$0.53\\
        		\midrule
    			Ours                               &\textbf{87.76$\pm$3.52} && \textbf{78.34$\pm$1.62} && \textbf{2.76$\pm$0.50} && \textbf{0.88$\pm$0.40}\\
        		\bottomrule 
		\end{tabular}}
	\end{center}
\end{table}

Table~\ref{tab2} presents the segmentation results on the NIH dataset, where our method consistently outperforms other state-of-the-art approaches across all metrics. In pancreas segmentation, FASS achieves a Dice score of 87.76\% and a Jaccard index of 78.34\%, indicating improved accuracy. For boundary accuracy, our method records 2.76 mm (HD95) and 0.88 mm (ASD), demonstrating a significant advantage over competing approaches.

\begin{figure*}[!t]
	\centering
	\includegraphics[width=\linewidth]{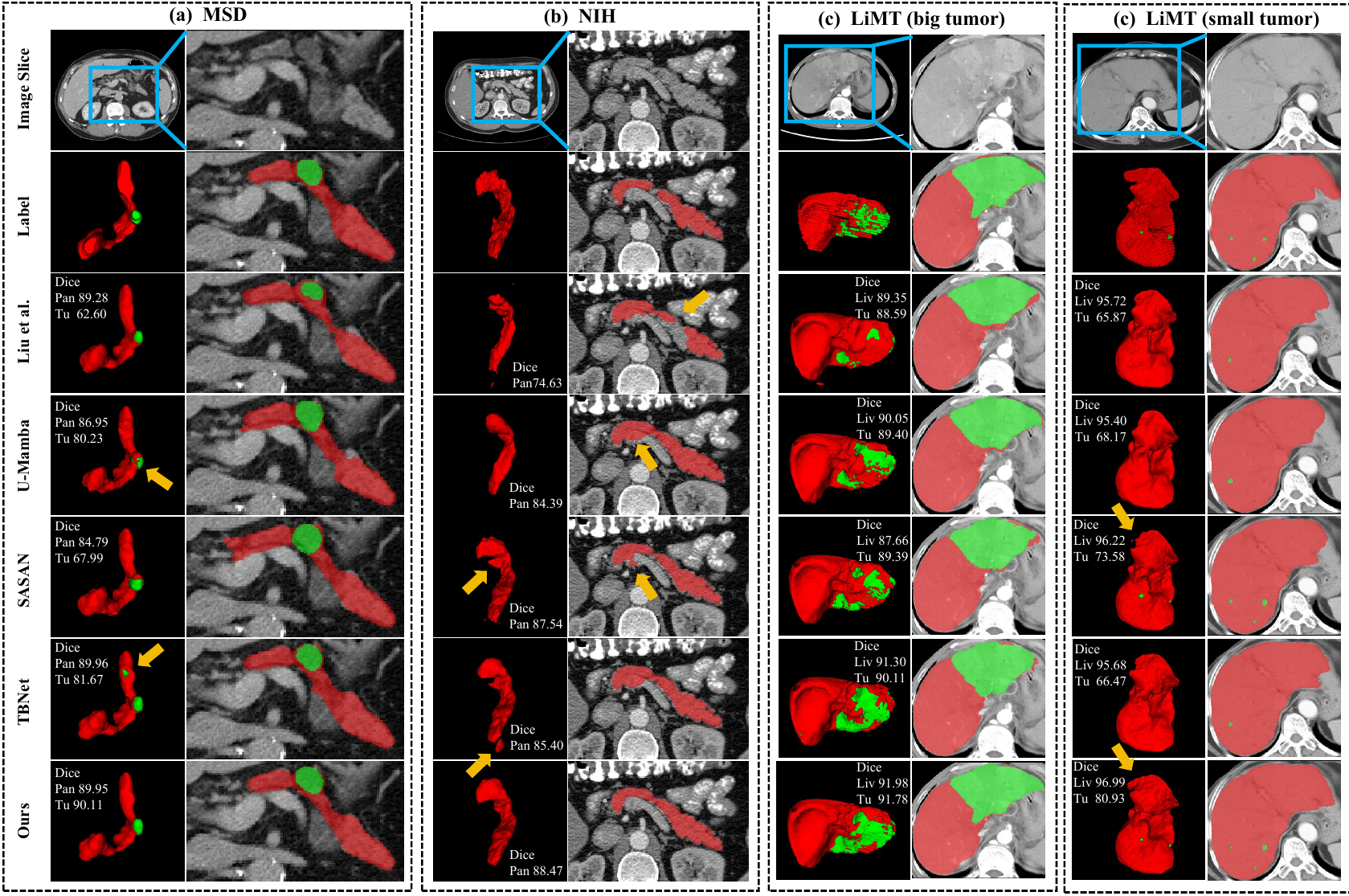}
 	\caption{Qualitative segmentation examples of our framework compared to competing approaches across three datasets demonstrate that our framework significantly enhances segmentation accuracy and integrity, especially in capturing complex tumor shapes and detecting small tumors.}
	\label{fig3}
\end{figure*}

Table~\ref{tab3} shows segmentation results on the LiMT dataset, with our method consistently achieving the highest scores across all metrics. In Dice and Jaccard scores, our method demonstrates substantial improvements in both liver and liver tumor segmentation. Specifically, it achieves a Dice score of 96.77\% and a Jaccard index of 93.79\% in liver segmentation, outperforming others by at least 1\%, underscoring its strength in liver region segmentation. For liver tumor segmentation, our model reaches a Dice score of 60.31\% and a Jaccard index of 43.47\%, exceeding frequency-domain methods like SASAN and TBNet by at least 2\%, reflecting improved precision in capturing tumor boundaries. Besides, our method achieves lower boundary error metrics, with a 95HD of 3.64 mm and an ASD of 1.01 mm for liver segmentation and 23.51 mm (95HD) and 5.64 mm (ASD) for tumor segmentation, confirming our model’s robustness in managing complex liver and tumor boundaries.

\subsubsection{Qualitative Analysis}

Fig.~\ref{fig3} shows a qualitative comparison of our framework with other methods. In the pancreas and small tumor co-segmentation example in Fig.~\ref{fig3} (a), U-Mamba fails to completely segment the tumor due to insufficient handling of the tumor-pancreas relationship; TBNet, with limited attention to global structure, results in false positive outputs (indicated by the yellow arrow); and RC-3DUNet and SASAN are affected by background tissues, resulting in missegmentation regions. In contrast, our framework focuses more effectively on the foreground region and considers the intrinsic connection between the organ and tumor. In Fig.~\ref{fig3}(b), the method by RC-3DUNet produces discontinuous pancreatic boundary segmentation with a Dice score of only 74.63\% (indicated by the yellow arrow), while our framework, with the introduction of the EC module, achieves finer and smoother segmentation, especially in the pancreatic head, closely matching the ground truth. Fig.~\ref{fig3}(c) presents a large primary liver cancer segmentation example from the LiMT dataset. Due to the tumor’s large size and distortion of liver morphology, RC-3DUNet, U-Mamba, and SASAN struggle to fully segment the tumor. Our framework achieves higher segmentation completeness, with Dice scores of 91.98\% and 91.78\% for liver and tumor, respectively. In the hemangioma segmentation example in Fig.~\ref{fig3}(d), despite the small tumor size, only SASAN and our framework successfully detect all tumors (indicated by the yellow arrow). In contrast, our framework closely aligns with the ground truth, achieving a Dice score of 96.99\% for the liver and 80.93\% for the tumor.

\begin{table*}
	\begin{center}
        \caption{Segmentation results on LiMT dataset compared with other state-of-the-art methods.}
		\label{tab3}
		\resizebox{\linewidth}{!}{
			\begin{tabular}{rccccccccccc}
				\toprule
				\multirow{3}{*}{Methods}&\multicolumn{10}{c}{Metrics}\\
				\cline{2-12}
				&\multicolumn{2}{c}{Dice [\%]$\uparrow$}&&\multicolumn{2}{c}{Jaccard [\%]$\uparrow$}&&\multicolumn{2}{c}{95HD [mm]$\downarrow$}&&\multicolumn{2}{c}{ASD [mm]$\downarrow$}\\
				\cline{2-3}\cline{5-6}\cline{8-9}\cline{11-12}
				&Liver&Tumor&&Liver& Tumor &&Liver&Tumor&&Liver&Tumor\\
				\midrule
				3D U-Net \cite{cciccek20163d}       &92.99$\pm$2.85  &53.84$\pm$5.25 && 86.94$\pm$2.27 & 36.87$\pm$5.03 && 8.13$\pm$5.17 & 37.92$\pm$10.83 && 3.41$\pm$1.73 & 9.34$\pm$4.76\\
				V-Net \cite{milletari2016v}         &91.56$\pm$3.17  &54.84$\pm$7.38 && 84.38$\pm$2.35 & 37.89$\pm$7.28 && 7.75$\pm$6.38 & 39.49$\pm$12.58 && 3.48$\pm$1.48 & 9.97$\pm$4.24\\
				Swin UNETR \cite{tang2022self}      &90.23$\pm$5.74  &50.52$\pm$7.65 && 82.49$\pm$4.63 & 33.97$\pm$7.03 && 6.75$\pm$5.02 & 31.98$\pm$9.33  && 4.23$\pm$1.80 & 8.76$\pm$4.45\\
                RC-3DUNet \cite{liu2024pancreas}   &90.17$\pm$3.58  &51.55$\pm$9.48 && 82.17$\pm$3.34 & 34.75$\pm$8.98 && 4.69$\pm$2.52 & 24.06$\pm$5.60  && 1.27$\pm$0.75 & 7.07$\pm$2.80\\
                nnU-Net (3D) \cite{isensee2021nnu}   &94.16$\pm$2.13  &57.19$\pm$4.36 && 88.97$\pm$2.26 & 40.15$\pm$4.01 && 4.54$\pm$2.59 & 27.77$\pm$5.00  && 1.29$\pm$0.61 & 6.43$\pm$2.42\\
                U-Mamba \cite{ma2024u}              &95.83$\pm$2.19  &56.65$\pm$3.86 && 92.19$\pm$1.98 & 39.56$\pm$3.50 && 4.31$\pm$1.43 & 25.78$\pm$4.96  && 1.46$\pm$0.79 & 6.62$\pm$2.49\\
        	\midrule                
				XU-Net \cite{li2020wavelet}         &94.35$\pm$2.18  &53.14$\pm$7.21 && 89.43$\pm$2.03 & 35.56$\pm$6.36 && 5.43$\pm$2.70 & 32.01$\pm$7.16  && 2.09$\pm$1.32 & 5.94$\pm$3.58\\
				XNet \cite{zhou2023xnet}            &92.73$\pm$3.02  &58.55$\pm$4.83 && 86.48$\pm$2.56 & 42.48$\pm$4.27 && 6.23$\pm$3.87 & 29.39$\pm$10.28 && 1.66$\pm$0.78 & 6.67$\pm$3.30\\
                FET \cite{azad2023unlocking}        &93.39$\pm$2.35  &56.64$\pm$4.02 && 87.98$\pm$1.75 & 39.58$\pm$3.81 && 4.83$\pm$1.67 & 26.93$\pm$7.88  && 1.83$\pm$1.05 & 6.31$\pm$2.43\\
                SASAN \cite{huang2024sasan}         &95.88$\pm$2.06  &58.81$\pm$3.88 && 92.06$\pm$1.75 & 41.69$\pm$3.26 && 3.98$\pm$2.01 & 24.46$\pm$5.36  && 1.08$\pm$0.67 & 5.73$\pm$3.13 \\
        	\midrule
				HMEDN \cite{zhou2019high}           &92.76$\pm$2.41  &53.37$\pm$8.46 && 86.75$\pm$1.87 & 36.57$\pm$5.89 && 4.27$\pm$1.50 & 25.58$\pm$6.03  && 1.20$\pm$0.73 & 5.82$\pm$2.82\\
				TBNet \cite{zhang2024low}           &95.83$\pm$1.76  &57.92$\pm$3.97 && 92.03$\pm$1.44 & 40.65$\pm$3.51 && 4.78$\pm$1.83 & 24.07$\pm$5.26  && 1.13$\pm$0.89 & 6.08$\pm$2.45\\
        	\midrule
				Ours                                &\textbf{96.77$\pm$1.46}  &\textbf{60.31$\pm$3.44} && \textbf{93.79$\pm$1.32} & \textbf{43.47$\pm$2.28} && \textbf{3.64$\pm$1.15} & \textbf{23.51$\pm$4.79}  && \textbf{1.01$\pm$0.59} & \textbf{5.64$\pm$2.21}\\
				\bottomrule 
		\end{tabular}}
	\end{center}
\end{table*}

\subsection{Ablation Study}
To evaluate the effectiveness of each module in FASS, we conducted detailed ablation experiments and selected 3D U-Net \cite{cciccek20163d} as the baseline network. The quantitative results summarized in Table~\ref{tab2} highlight the significant performance improvements contributed by each module. Among them, the FA module demonstrated the core value in the pancreas segmentation task on the MSD and NIH datasets, effectively reducing the false positive prediction of the background region. This improvement is attributed to the model's ability to accurately focus on pancreas and tumors while neighboring tissues have similar grayscale values with them. However, on the LiMT dataset, where the surrounding tissues of the liver are relatively less similar and the liver shape is more regular, the FA module showed minimal improvement in liver segmentation accuracy. Its advantage became more evident in liver tumor segmentation. The FLFE module performed exceptionally well in liver tumor segmentation on the LiMT dataset. Thanks to the enhancement of features in the frequency domain, different types of liver tumors are more discriminative, thereby optimizing the discriminant ability of the model. The EC module showed limited boundary constraint effects on the NIH dataset, possibly due to the annotation quality, as the unsmooth boundary of ground truth in the NIH dataset may limit the effective application of edge constraints. Furthermore, the combination of different modules can enhance the model's performance to varying degrees. In summary, the experimental results show that the collaborative use of the FA, FLFE, and EC modules leads to superior segmentation performance.

\begin{table*}
	\begin{center}
		\caption{Ablation study of each module on three datasets.}
		\label{tab4}
		\resizebox{0.95\linewidth}{!}{
			\begin{tabular}{cccccccccccc}
				\toprule
				\multicolumn{4}{c}{Methods} & & \multicolumn{2}{c}{MSD dataset} & & NIH dataset  & & \multicolumn{2}{c}{LiMT dataset} \\
				\cline{1-4} \cline{6-7} \cline{9-9} \cline{11-12}
				 Baseline & FA & FLFE & EC & & Pancreas & Tumor & & Pancreas &  & Liver & Tumor \\
				\midrule
				\checkmark & & & && 79.52$\pm$7.54 & 43.18$\pm$28.50 & & 79.48$\pm$5.06 && 92.99$\pm$2.85 & 53.84$\pm$5.25\\
    			\checkmark & \checkmark & & && 83.02$\pm$5.44 & 56.47$\pm$20.23 && 83.57$\pm$4.38 && 93.07$\pm$2.61 & 57.18$\pm$4.07\\
       			\checkmark & &\checkmark &  && 82.77$\pm$5.97 & 49.37$\pm$22.79 && 82.23$\pm$4.82 && 94.18$\pm$1.93 & 57.21$\pm$3.89\\
                    \checkmark &  & &\checkmark && 81.03$\pm$6.67 & 46.29$\pm$27.91 && 80.04$\pm$5.36 && 94.60$\pm$2.75 & 55.61$\pm$4.47\\         
                    \checkmark & \checkmark & & \checkmark && 85.36$\pm5.72$ & 56.95$\pm$22.57 && 85.15$\pm$2.37 && 95.42$\pm$2.53 & 58.48$\pm$3.81\\
                    \checkmark & & \checkmark & \checkmark && 84.91$\pm$5.98 & 52.73$\pm$25.41 && 83.44$\pm$4.93 && 95.82$\pm$2.44 & 58.56$\pm$3.56\\
                    \checkmark & \checkmark & \checkmark & && 86.73$\pm$5.85 & 58.36$\pm$19.55 && 86.61$\pm$3.89 && 96.23$\pm$1.57 & 59.28$\pm$3.72\\
				\checkmark & \checkmark & \checkmark & \checkmark && \textbf{87.85$\pm$5.30} & \textbf{60.49$\pm$18.26} && \textbf{87.76$\pm$3.52}&& \textbf{96.77$\pm$1.46} & \textbf{60.31$\pm$3.44}\\
				\bottomrule 
		\end{tabular}}
	\end{center}
\end{table*}

\subsubsection{FA Module}
\begin{figure}[!t]
	\centering
	\includegraphics[width=0.95\columnwidth]{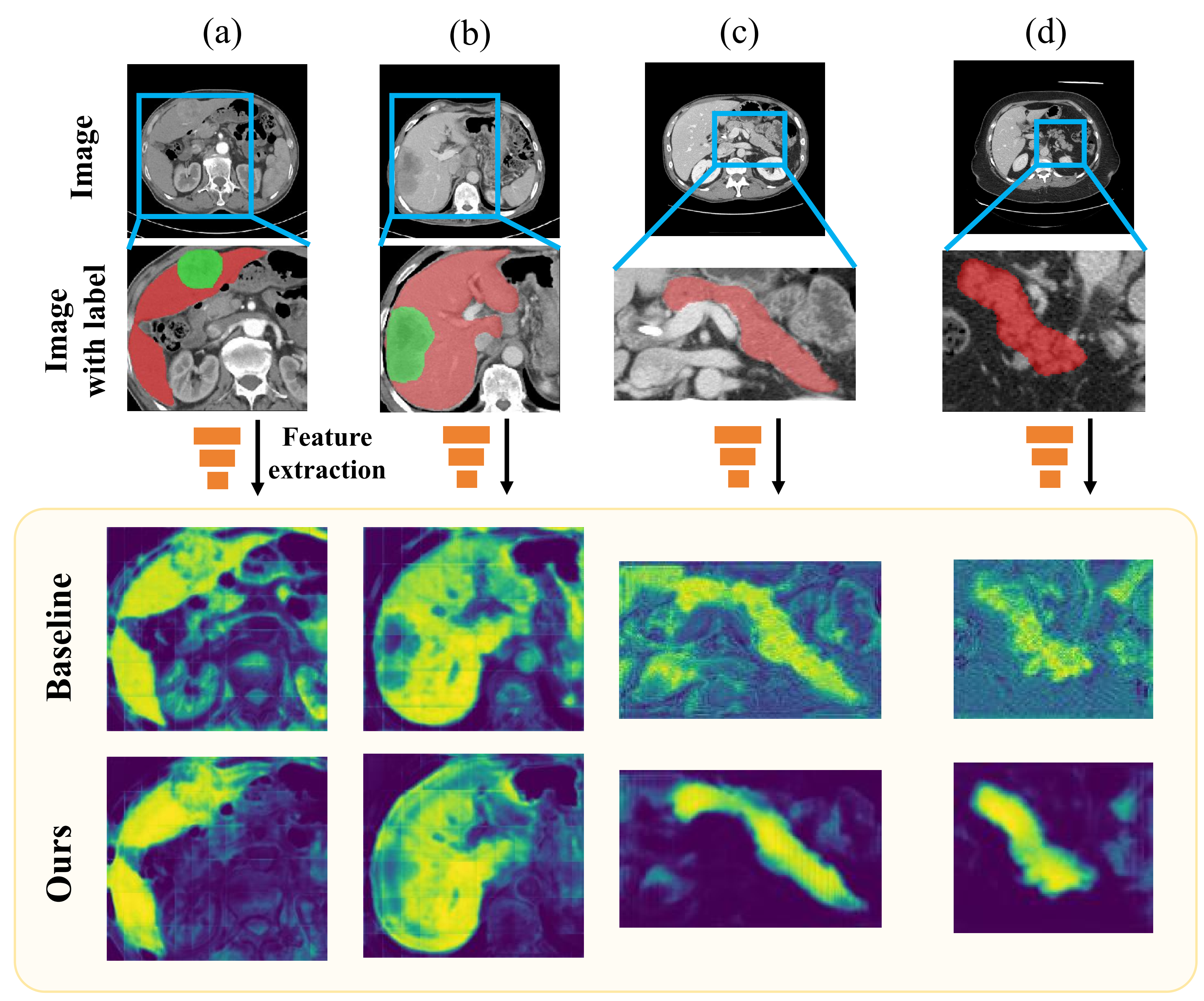}
 	\caption{Visual analysis of encoder feature extraction. With the introduction of the FA module, our method effectively focuses on foreground areas, filtering out the complex background information.}
	\label{fig4}
\end{figure}

To evaluate the effectiveness of the FA module, we performed a visual analysis of the features extracted by the encoder layer during the inference stage. Fig.~\ref{fig4} shows the performance of the module's ability to focus on foreground regions on the LiMT dataset and the MSD pancreas dataset. Pancreas and liver are usually surrounded by complex backgrounds, resulting in low contrast at the boundaries with surrounding tissue. Compared with the baseline method, the introduction of the FA module led to greater unbiased attention to the foreground features. Specifically, Fig.~\ref{fig4} (a) and Fig.~\ref{fig4} (b) demonstrate the feature extraction of liver images with tumors within a complex scenario. In this scenario, the features extracted by the baseline model contain irrelevant background information, resulting in cluttered and unfocused representations. In contrast, when the FA module is introduced, the model successfully separates the foreground features from the entire image features, effectively filtering out the complex background information.

\begin{figure}[!t]
	\centering
	\includegraphics[width=0.95\columnwidth]{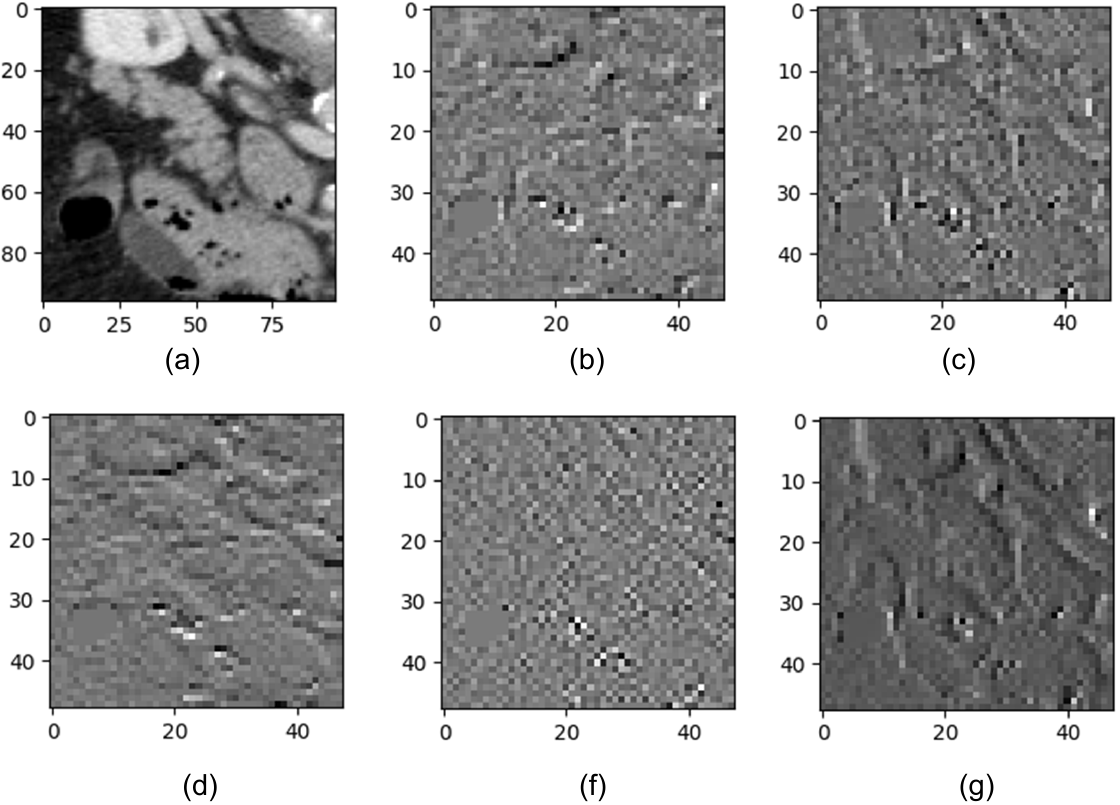}
 	\caption{Visualization comparison of high-frequency features: (a) shows the original image, (b) displays the feature map extracted after the separable convolution operation, and (c)-(f) present the high-frequency components in the vertical, horizontal, and diagonal directions, respectively. (g) illustrates the feature map after integrating each component through the cross-attention mechanism.}
	\label{fig5}
\end{figure}

Table.~\ref{tab5} reports the performance and efficiency of the algorithm under different background sampling sizes. The results show that when the sampling size is set to 18$\times$18$\times$18, the model tends to focus excessively on smaller local background areas, which may limit its learning ability and, in turn, affect segmentation performance. In contrast, when the sampling size is set to 48$\times$48$\times$48, it becomes more difficult to select background areas with low overlap with the foreground regions, hindering the design goals of the foreground perception module and increasing sampling time. In comparison, a sampling size of 32$\times$32$\times$32 allows the model to efficiently sample appropriate background areas in a shorter time, demonstrating the best segmentation performance across the three datasets.

\begin{table}
	\begin{center}
		\caption{Comparison of performance and time for different background region sampling sizes on the MSD dataset.}
		\label{tab5}
		\resizebox{0.9\linewidth}{!}{
			\begin{tabular}{cccc}
				\toprule
				\multirow{2}{*}{Size} & \multicolumn{2}{c}{Dice (\%) $\uparrow$}  & \multirow{2}{*}{Sampling time (ms) $\downarrow$}\\
				\cline{2-3}
				& Pancrea & Tumor & \\
				\midrule
				18$\times$18$\times$18 & 84.78$\pm$7.35 & 57.15$\pm$20.06 & \textbf{1.67}\\
				32$\times$32$\times$32 & \textbf{87.85$\pm$5.30} & \textbf{60.49$\pm$18.26} & 2.18\\
				48$\times$48$\times$48 & 83.02$\pm$6.64 & 58.30$\pm$21.23 & 8.54\\
				\bottomrule 
		\end{tabular}}
	\end{center}
\end{table}

\begin{figure}[!t]
	\centering
	\includegraphics[width=0.95\columnwidth]{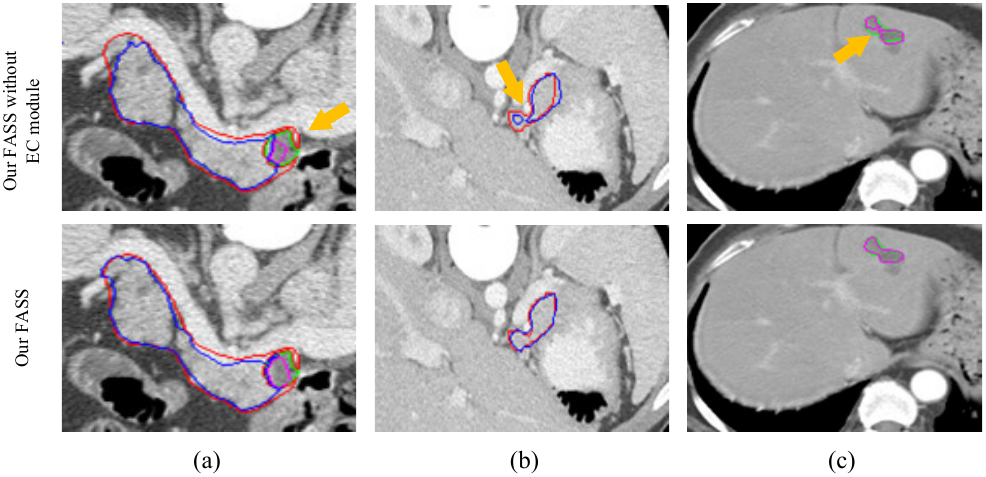}
 	\caption{Visualization of example results with and without the edge constraint module. (a), (b), and (c) show sample results from the MSD pancreas dataset, the NIH dataset, and the LiMT dataset, respectively. Red and green lines represent the ground truth for organs and tumors, while blue and purple lines indicate the predicted boundaries for organs and tumors. With the introduction of the EC module, our method achieves improved boundary continuity and smoothness.}
	\label{fig6}
\end{figure}

\subsubsection{FLFE Module}
In medical images, the precise capture and utilization of high-frequency features are crucial for detail recognition and edge delineation. The proposed method enhances the representation of high-frequency details by applying cross-attention among high-frequency components in the horizontal, vertical, and diagonal directions, effectively leveraging their complementary advantages. As shown in Fig.~\ref{fig5}(b), experiments on the MSD pancreas dataset demonstrate that although the direct convolution operations can extract basic features, the edge and texture performance are slightly blurred. In contrast, as shown in Fig.~\ref{fig5}(g), after integrating high-frequency components through the cross-attention mechanism, the resulting feature map exhibits clearer edges and texture details. This demonstrates the effective complementarity of high-frequency components from Fig.~\ref{fig5}(c) to Fig.~\ref{fig5}(f).

\begin{figure*}[!t]
\centering
\includegraphics[width=\linewidth]{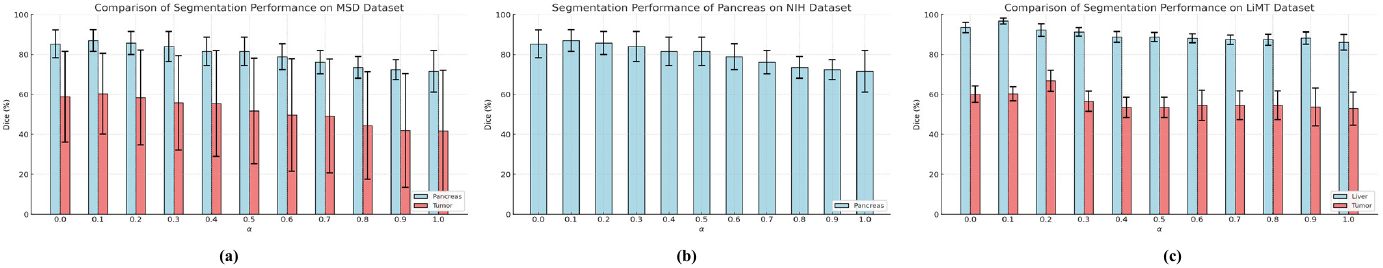}
\caption{Comparison of the impact of different $\alpha$ values on model performance. Where (a), (b), and (c) represent the Dice scores and the corresponding standard deviations on the MSD pancreas dataset, NIH dataset, and LiMT dataset, respectively.}
\label{fig7}
\end{figure*}

The selection of different wavelet bases affects the performance and time efficiency of the experiment. Therefore, we compare the application effects of various wavelet bases on the MSD pancreas dataset, including Haar, Biorthogonal (Bior 2.4), Coiflets (Coif 1), and Daubechies (Db 2), as shown in Table~\ref{tab6}. The experimental results show that although the Haar wavelet basis exhibits high boundary sensitivity, due to its simple structure, its segmentation accuracy is limited in dealing with texture details, particularly in tumor segmentation. The Coif 1 wavelet basis achieves the best performance in tumor segmentation due to its excellent symmetry and attenuation characteristics, but this comes at the cost of high computational complexity and training time. In contrast, the Db 2 wavelet basis maintains both high segmentation accuracy and time efficiency. Therefore, Db 2 is chosen as the benchmark wavelet basis for the wavelet transform in this paper.

\begin{table*}
	\begin{center}
		\caption{Comparison of different wavelet bases on MSD dataset.}
		\label{tab6}
		\resizebox{0.8\linewidth}{!}{
			\begin{tabular}{ccccccc}
				\toprule
				\multirow{2}{*}{Wavelet} & \multicolumn{2}{c}{Dice (\%) $\uparrow$} & & \multicolumn{2}{c}{ASD (mm) $\downarrow$} & \multirow{2}{*}{\makecell{Wavelet decomposition\\ time (ms) $\downarrow$}} \\
				\cline{2-3}\cline{5-6}
				& Pancreas &  Tumor & & Pancreas & Tumor & \\
				\midrule
				Haar & 84.76$\pm$6.33 & 58.96$\pm$22.82  & & 1.65$\pm$0.81 & 7.72$\pm$2.08 & \textbf{0.26}\\
				Bior 2.4 & 84.13$\pm$6.94 & 57.44$\pm$21.76 & & 1.66$\pm$0.65 & 11.09$\pm$2.53 & 1.06 \\ 
				Coif 1 & 86.59$\pm$5.27 & \textbf{60.78$\pm$20.54} & & 1.41$\pm$0.34 & 6.12$\pm$1.11 & 2.31\\
				Db 2 & \textbf{87.85$\pm$5.30} & 60.49$\pm$18.26 & & \textbf{1.06$\pm$0.21} & \textbf{4.55$\pm$0.87} & 0.73\\
				\bottomrule 
		\end{tabular}}
	\end{center}
\end{table*}

\subsubsection{EC Module}

To further demonstrate the effectiveness of the EC module, visual results of selected samples are presented, as shown in Fig.~\ref{fig6}. The baseline segmentation results (see the first row of Fig.~\ref{fig6}) exhibit poor boundary continuity and smoothness, particularly in the areas indicated by the yellow arrows. In contrast, after incorporating the EC module, the segmentation boundaries in the second row of Fig.~\ref{fig6} show significant improvement. This change indicates that the EC module plays a crucial role in enhancing the segmentation results, improving the model's adaptability to low-contrast images, and consequently increasing overall segmentation accuracy.

\subsection{Parameter Sensitivity Analysis}\label{sec4.5}
\textbf{Impact of parameter $\alpha$:} As a key regulatory factor in the FA module, the parameter $\alpha$ dominates the sampling position of the background region and ranges from [0, 1]. As shown in Fig.~\ref{fig7}, this experiment systematically explored the specific impact of different $\alpha$ values on model performance. Theoretically, an $\alpha$ value closer to 1 indicates fewer background elements in the sampling region. Ideally, $\alpha=0$ represents an ideal adversarial training pattern that is sampled entirely from the background. However, the experimental results show that the peak performance is not achieved when $\alpha=0$ but rather in a small range near zero. Specifically, when $\alpha$ is set to a relatively low value (e.g., 0.1), the model shows excellent performance on three datasets. In this case, the selected background region is close to the foreground boundary with minimal overlap, but it still moderately expands the background features compared to $\alpha>0.1$. This phenomenon may be due to the fact that a small number of overlaps will deepen the model's understanding of low-contrast adjacent background tissues. On the contrary, a higher $\alpha$ value results in excessive inclusion of foreground information during sampling, contrary to the design intent of the FA module, thus negatively affecting its efficacy. When $\alpha$ is set to 0.2, the model shows slight improvement in tumor segmentation performance on the LiMT dataset compared to $\alpha=0.1$. This may be due to the model incorrectly identifying liver tissue as part of the background, unintentionally reducing phenotypic contrast between the liver and tumors, which, in turn, stimulates the model to pay more attention to the tumor region, thereby improving the segmentation accuracy to some extent. Based on the analysis of the experimental results of parameter $\alpha$, when $\alpha=0.1$, the model shows excellent segmentation performance on the three datasets, which confirms the importance of reasonable choice of $\alpha$ value for the overall segmentation performance.

\section{Conclusion}
In this paper, we proposed a foreground-aware spectrum segmentation (FASS) framework for low-contrast medical images. First, the foreground-aware module forces the model to focus on the target areas through adversarial training of background features and global features. Then, a feature-level frequency enhancement strategy is designed to better segment fine anatomical structures, along with an edge constraint that aligns edge prediction with expected contours, enhancing boundary continuity. Extensive experiments demonstrate that the proposed method outperforms others in segmentation performance across multiple medical image datasets. FASS not only holds promise for advancing the clinical application of low-contrast medical image segmentation but also provides more reliable data support for clinical decision-making.

%\section{References Section}
%You can use a bibliography generated by BibTeX as a .bbl file.
% BibTeX documentation can be easily obtained at:
% http://mirror.ctan.org/biblio/bibtex/contrib/doc/
% The IEEEtran BibTeX style support page is:
% http://www.michaelshell.org/tex/ieeetran/bibtex/

\bibliographystyle{IEEEtran}
\bibliography{mybibliography}

\end{document}